\documentclass[conference]{IEEEtran}
\usepackage{amsmath,amssymb}
\usepackage{booktabs}
\usepackage{graphicx}
\usepackage{hyperref}
\usepackage{cite}
\usepackage{multirow}
\usepackage{xcolor}
\usepackage{float}
\usepackage{enumitem}

\hypersetup{colorlinks=true,linkcolor=blue,citecolor=blue,urlcolor=blue}

\begin{document}

\title{AXELRAM: Quantize Once, Never Dequantize}

\author{\IEEEauthorblockN{Yasushi Nishida}
\IEEEauthorblockA{Axelidea Inc.\\
\url{https://axelidea.com}}}

\maketitle

\begin{abstract}
We propose AXELRAM, a smart SRAM macro architecture that computes attention scores directly from quantized KV cache indices without dequantization.
The key enabler is a \emph{design-time fixed codebook}: orthogonal-transform-based quantization concentrates each coordinate's distribution to $\mathcal{N}(0,1/d)$, so the optimal quantizer depends only on dimension~$d$ and bit-width~$b$, not on input data.
The asymmetric path design---transform on write, table-lookup on read with no inverse transform---reduces per-query multiplications by $102.4\times$ (a mathematical identity).

Through multi-seed evaluation (10 seeds $\times$ 3 models), we discover that \emph{sign pattern sensitivity} causes catastrophic PPL spikes ($\Delta > 50$) on certain models (Qwen2.5-3B), while others (LLaMA-3.1-8B) are fully stable.
This phenomenon extends SpinQuant's observation of rotation variance in weight quantization to the KV cache domain, where the effect is qualitatively more severe.
We trace the root cause to layer-wise norm heterogeneity and propose a gradient-free sign pattern selection (200 candidates, 8 calibration samples, one-time) that eliminates catastrophic spikes with zero additional hardware cost.
All source code is available at \url{https://github.com/Axelidea/AXELRAM}.
\end{abstract}

\section{Introduction}

Large language model (LLM) inference is bottlenecked by key-value (KV) cache memory: an 8B-parameter model at context length $T{=}4096$ stores 512~MB in FP16.
KV cache quantization to 2--4 bits reduces storage, but the quantized keys must be \emph{dequantized} before computing attention scores---$T$ times per query.

We ask: \textbf{can the SRAM that stores the quantized cache also compute attention scores, eliminating dequantization entirely?}

\begin{figure}[t]
\centering
\includegraphics[width=\columnwidth]{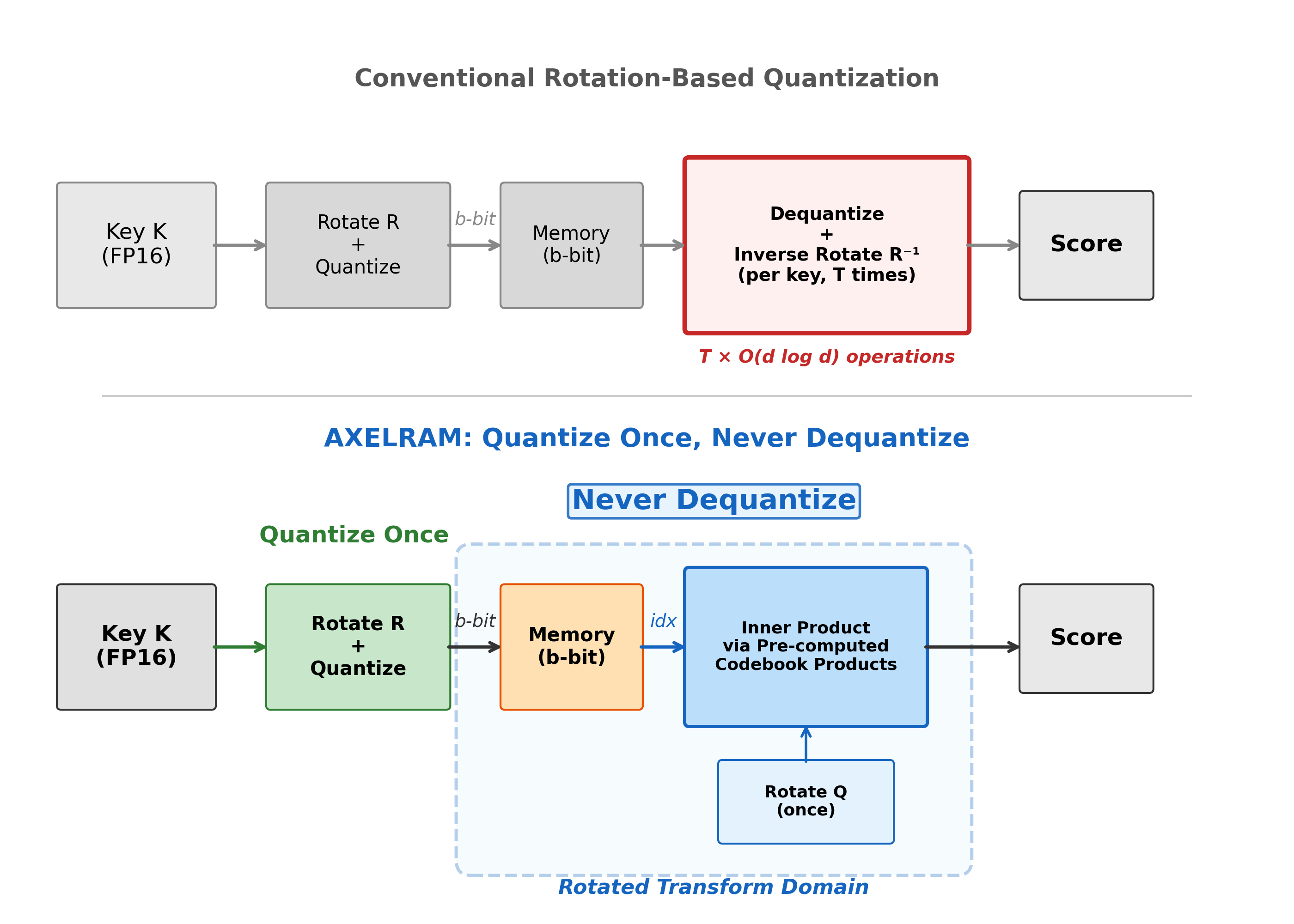}
\caption{Quantize Once, Never Dequantize. Conventional rotation-based quantization (top) must dequantize and inverse-rotate each stored key before computing attention---$T \times O(d\log d)$ per query. AXELRAM (bottom) computes attention entirely within the rotated transform domain: the query is rotated once, and scores are computed via pre-computed codebook products with no dequantization. The orthogonal invariance $\langle \mathbf{q}, \mathbf{k} \rangle = \langle R\mathbf{q}, R\mathbf{k} \rangle$ makes this possible, yielding $102.4\times$ fewer multiplications.}
\label{fig:comparison}
\end{figure}

AXELRAM answers this by integrating orthogonal-transform-based quantization with table-lookup attention in a single SRAM macro (Fig.~\ref{fig:comparison}).
The design-time fixed codebook (30 bytes for $b{=}3$) is shared across all $d$ dimensions and stored in ROM.

\textbf{Contributions.}
\begin{enumerate}[leftmargin=*,nosep]
\item We propose AXELRAM, a smart SRAM macro with an asymmetric write/read path that reduces multiplications by $102.4\times$ (Section~\ref{sec:arch}).
\item We discover \emph{sign pattern sensitivity} in rotation-based KV cache quantization through 10-seed evaluation across 3 models, finding catastrophic spikes ($\Delta > 50$) unreported in prior work (Section~\ref{sec:sensitivity}).
\item We propose a gradient-free sign pattern selection that eliminates spikes with \emph{zero additional hardware cost} (Section~\ref{sec:mitigation}).
\end{enumerate}

\section{Principle: Why the Codebook Is Fixed}
\label{sec:principle}

For unit-norm $\hat{\mathbf{x}} \in \mathbb{R}^d$ and orthogonal $R$, each coordinate of $\mathbf{y} = R\hat{\mathbf{x}}$ follows approximately $\mathcal{N}(0,1/d)$ for $d \geq 64$ (concentration of measure~\cite{turboquant}).
The Lloyd-Max quantizer for this distribution yields centroids and boundaries as functions of $(d,b)$ alone---a \textbf{design-time fixed codebook}.
For $b{=}3$, $d{=}128$: 8~centroids + 7~boundaries = 15~FP16 values = 30~bytes, verified to satisfy the Lloyd-Max optimality conditions to machine precision ($< 10^{-11}$).

The inner product is preserved under orthogonal transforms: $\mathbf{q}^\top \mathbf{k} = (R\mathbf{q})^\top(R\mathbf{k})$.
Instead of inverse-transforming each stored key ($T$ times), we transform the query once and compute in the rotated transform domain:
\begin{equation}
\langle \mathbf{q}, \hat{\mathbf{k}} \rangle
= \langle R\mathbf{q},\; \mathrm{codebook}[\mathrm{idx}] \rangle \times \|\mathbf{k}\|
\label{eq:asymmetric}
\end{equation}
This is the \emph{asymmetric path} that enables table-lookup attention.

\textbf{Contrast with TurboQuant.}
TurboQuant~\cite{turboquant} computes $\langle \mathbf{q},\; R^{-1} \cdot \mathrm{codebook}[\mathrm{idx}] \cdot \|\mathbf{k}\| \rangle$, applying inverse rotation to each key.
Equation~\eqref{eq:asymmetric} moves the rotation from keys ($T$ times) to the query (once), enabling the pre-computation table.
This rearrangement is not disclosed or suggested in~\cite{turboquant}.

\section{Architecture: AXELRAM Smart SRAM Macro}
\label{sec:arch}

\begin{figure*}[t]
\centering
\includegraphics[width=0.85\textwidth]{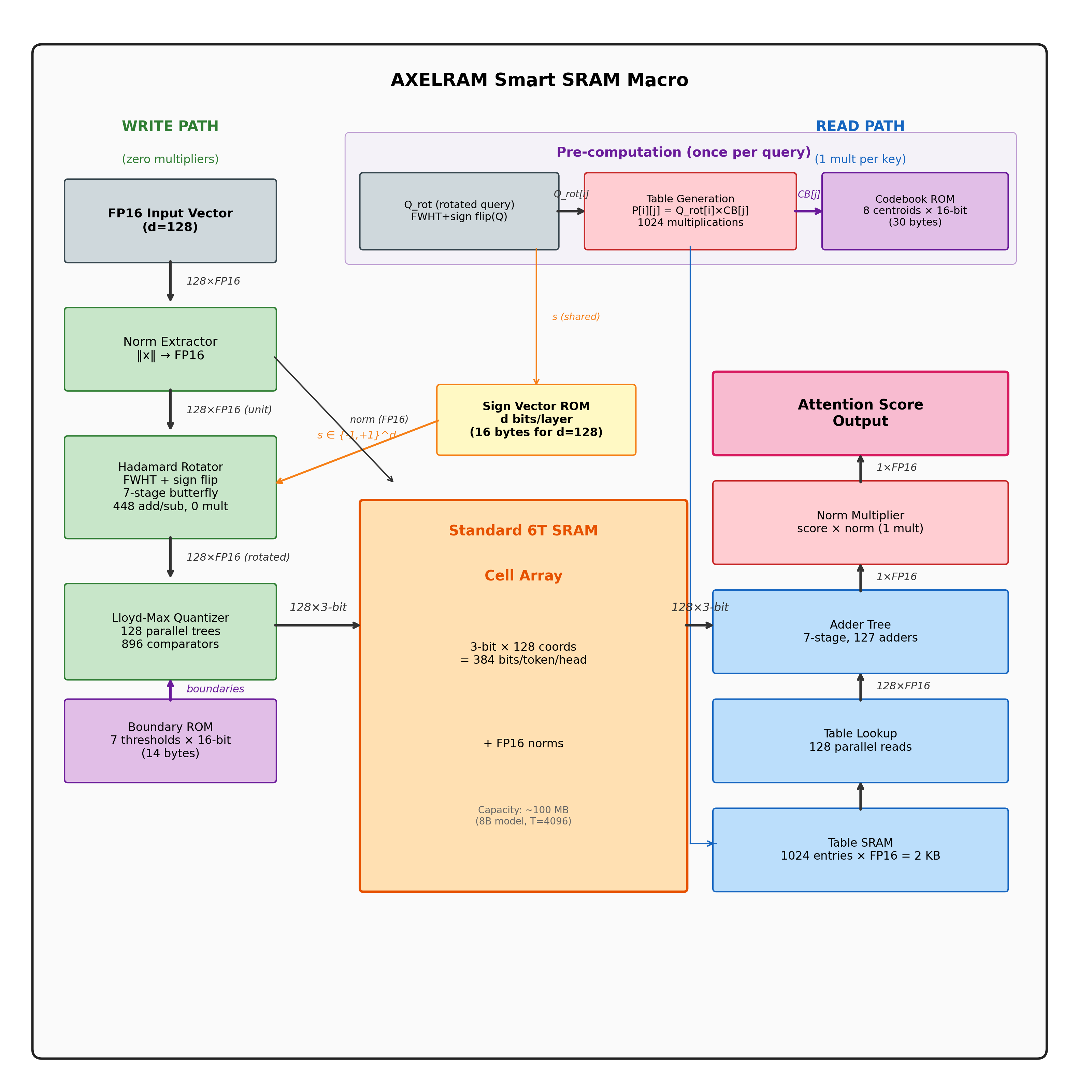}
\caption{AXELRAM smart SRAM macro. \textbf{Write path} (left): norm extraction, FWHT butterfly network (448 add/sub, zero multipliers), Lloyd-Max comparator quantization (896 comparators), writing 3-bit indices + FP16 norm to SRAM. \textbf{Read path} (right): pre-computed table lookup (128 parallel reads), adder tree (127 adders), norm scaling (1 multiplication). \textbf{Pre-computation} (top center, once per query): table generation with 1024 multiplications. The fixed codebook ROM (30 bytes) is shared by both paths across all $d{=}128$ dimensions.}
\label{fig:macro}
\end{figure*}

\subsection{Write Path (Zero Multipliers)}
The write path (Fig.~\ref{fig:macro}, left) applies a randomized Hadamard transform $R = H_d \cdot \mathrm{diag}(\mathbf{s})$ via a butterfly adder network (FWHT, $\log_2 d$ stages, $d/2$ add/sub per stage, zero multipliers), followed by comparator-based quantization against fixed codebook boundaries stored in ROM.
The sign vector $\mathbf{s} \in \{-1,+1\}^d$ is stored in ROM/eFuse (16 bytes for $d{=}128$).

\subsection{Read Path (Table Lookup + Accumulate)}
\begin{figure*}[t]
\centering
\includegraphics[width=\textwidth]{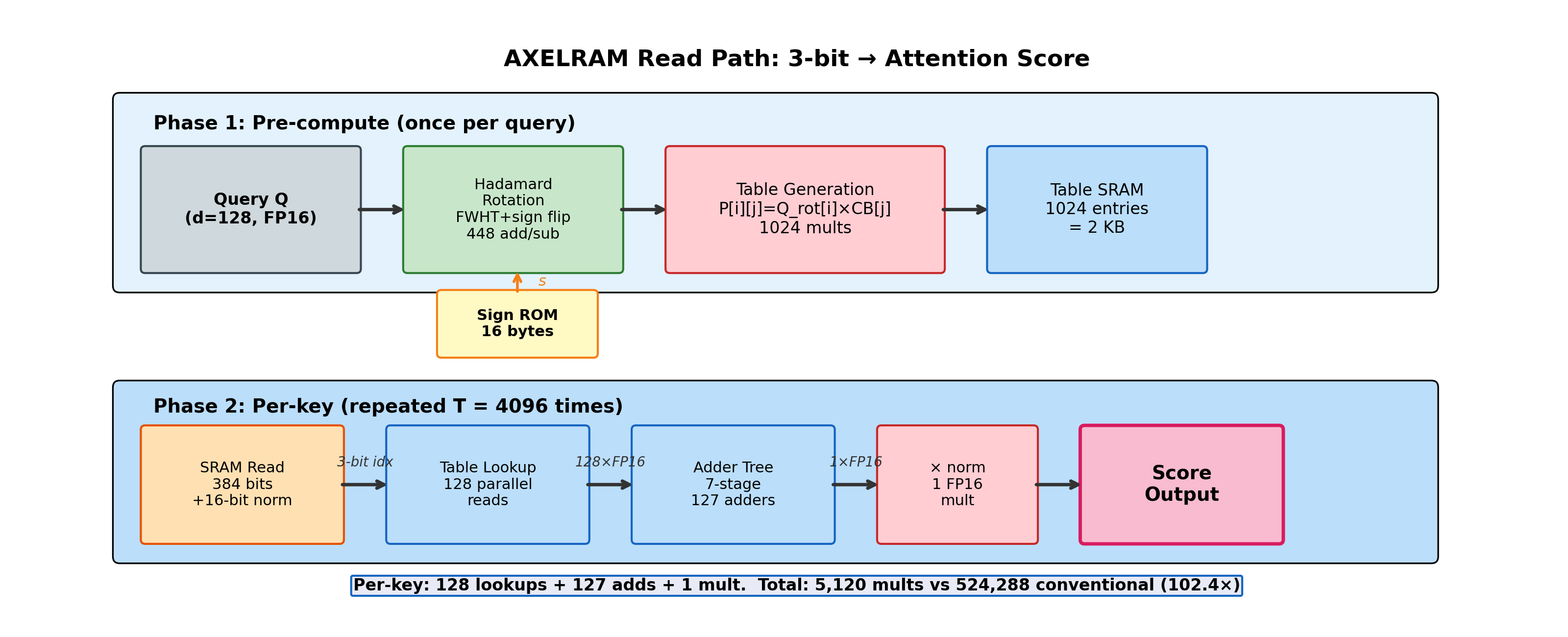}
\caption{Read path detail. Phase~1 (once per query): FWHT rotation of query, table generation ($d \times 2^b$ multiplications). Phase~2 (repeated $T$ times): table lookup, adder tree, norm scaling (1 multiplication per key). Total: 5,120 multiplications for $T{=}4096$ versus 524,288 conventional ($102.4\times$ reduction).}
\label{fig:readpath}
\end{figure*}

The read path (Fig.~\ref{fig:macro}, right; detailed in Fig.~\ref{fig:readpath}) operates in two phases.
Upon query arrival:
\begin{enumerate}[leftmargin=*,nosep]
\item Transform query: $\mathbf{q}_\mathrm{rot} = R \mathbf{q}$ (one FWHT, same circuit as write path).
\item Pre-compute table: $P[i][j] = q_{\mathrm{rot},i} \times c_j$ for $i{=}0,\ldots,d{-}1$, $j{=}0,\ldots,2^b{-}1$ ($d \times 2^b$ multiplications, once per query).
\item Per-key: read $d$ indices from SRAM, look up $P[i][\mathrm{idx}_i]$, accumulate via adder tree ($d{-}1$ additions), multiply by norm (1 multiplication).
\end{enumerate}

The read path contains \emph{no inverse transform circuit}.
The inner product is computed entirely from the transformed query coordinates, codebook centroids, and stored indices.

\subsection{Multiplication Count}
\begin{equation}
\underbrace{d \cdot 2^b}_{\text{table}} + \underbrace{T \cdot 1}_{\text{norms}}
= 1024 + 4096 = 5{,}120
\end{equation}
versus conventional $T \cdot d = 524{,}288$: a $\mathbf{102.4\times}$ reduction.
This is a mathematical identity derived from the inner-product preservation of orthogonal transforms, not an approximation.

\subsection{Memory}
Fixed codebook ROM: 30~bytes ($b{=}3$).
Sign vector ROM: 16~bytes/layer.
Pre-computation table SRAM: 2~KB.
KV cache: 50~bytes/vector (vs.\ 256~bytes FP16), $5.1\times$ compression.
For an 8B model with 32 layers and 8 KV heads at $T{=}4096$: approximately 100~MB (vs.\ 512~MB in FP16).

\section{Sign Pattern Sensitivity}
\label{sec:sensitivity}

\subsection{Background: SpinQuant}
SpinQuant~\cite{spinquant} demonstrated that random rotation matrices induce significant performance variance in \emph{weight-activation quantization} (W4A4), with up to 6-point accuracy spread across 100 random Hadamard rotations on LLaMA-2-7B.
They propose Cayley-parameterized gradient optimization to learn a fixed rotation.
We investigate whether analogous sensitivity exists in \emph{KV cache quantization}, which operates in a fundamentally different regime: online quantization of activation vectors rather than offline weight quantization.

\subsection{Experimental Setup}
We evaluate Hadamard rotation with random sign patterns $\mathbf{s} \in \{-1,+1\}^d$ across 10 seeds ($\{1, 2, \ldots, 10\}$) on three models:
LLaMA-3.1-8B-Instruct ($d{=}128$, 32 layers),
Qwen2.5-3B-Instruct ($d{=}128$, 36 layers), and
Qwen3-8B ($d{=}128$, 36 layers).
Perplexity is measured on WikiText-2 test set (stride 512, max length 2048).
Models are loaded with 4-bit NF4 weight quantization; KV cache quantization is the variable under study.
All source code and results are available at \url{https://github.com/Axelidea/AXELRAM}.

\subsection{Results}

\begin{table}[t]
\centering
\caption{Sign pattern sensitivity: PPL increase ($\Delta$) over FP16 baseline. Mean $\pm$ std over 10 seeds. \emph{Worst} = maximum $\Delta$ observed.}
\label{tab:sensitivity}
\setlength{\tabcolsep}{4pt}
\begin{tabular}{llccc}
\toprule
\textbf{Model} & & \textbf{2-bit} & \textbf{3-bit} & \textbf{4-bit} \\
\midrule
\multirow{2}{*}{\shortstack[l]{LLaMA-3.1-8B\\(FP16: 6.98)}}
& mean$\pm$std & +0.10$\pm$0.01 & +0.02$\pm$0.00 & +0.00$\pm$0.00 \\
& worst & +0.11 & +0.02 & +0.01 \\
\midrule
\multirow{2}{*}{\shortstack[l]{Qwen2.5-3B\\(FP16: 8.29)}}
& mean$\pm$std & +8.13$\pm$16.9 & +9.64$\pm$14.9 & +1.10$\pm$0.85 \\
& worst & \textbf{+58.4} & \textbf{+51.0} & +2.98 \\
\midrule
\multirow{2}{*}{\shortstack[l]{Qwen3-8B\\(FP16: 9.07)}}
& mean$\pm$std & +12.2$\pm$2.51 & +0.48$\pm$0.29 & +0.04$\pm$0.01 \\
& worst & +17.5 & +1.09 & +0.05 \\
\bottomrule
\end{tabular}
\end{table}

Table~\ref{tab:sensitivity} reveals three findings:

\textbf{Finding 1: Catastrophic spikes.}
On Qwen2.5-3B, certain sign patterns cause $\Delta > 50$---an order of magnitude worse than SpinQuant's 6-point variance in W4A4.
This catastrophic behavior is unreported in any prior KV cache quantization work, including TurboQuant~\cite{turboquant}, which reports single-seed results.

\textbf{Finding 2: Model dependency.}
LLaMA-3.1-8B is fully stable (std $< 0.01$ at all bit-widths), while Qwen2.5-3B exhibits catastrophic variance (std $> 14$).
The difference correlates with layer-wise norm heterogeneity:
Qwen2.5-3B layer~0 has a mean key norm of 172, versus 22 at other layers ($7.8\times$ ratio); LLaMA-3.1-8B shows a ratio of approximately $1.5\times$.
A sign pattern that concentrates high-norm channels' energy into few post-rotation coordinates violates the $\mathcal{N}(0,1/d)$ assumption, causing the fixed codebook to fail catastrophically.

\textbf{Finding 3: Bit-width interaction.}
The bit-width at which the spike occurs varies by seed.
On Qwen2.5-3B, seed~1 spikes at 2-bit, seed~3 at 3-bit, and seed~5 at 4-bit.
This rules out a simple ``low-bit-width problem'' explanation and suggests a complex interaction between the sign pattern and the model's layer-specific weight structure.

\subsection{Why QJL Hurts at Low Bit-Widths}
TurboQuant~\cite{turboquant} employs QJL~\cite{qjl} (Quantized Johnson-Lindenstrauss) 1-bit residual correction for unbiased inner-product estimation, ``stealing'' 1 bit from the MSE codebook.
Our multi-seed data reveals that QJL is harmful at $b \leq 3$:
the bit-stealing penalty (reducing MSE resolution from $b$ to $b{-}1$ bits) outweighs the bias correction.

\textbf{Theoretical explanation.}
The QJL estimator is unbiased for the \emph{inner product} $\langle \mathbf{q}, \mathbf{k} \rangle$, but attention requires $\mathrm{softmax}(\cdot)$, which applies $\exp(\cdot)$.
By Jensen's inequality, for convex $\exp$:
\begin{equation}
\mathbb{E}[\exp(X)] \geq \exp(\mathbb{E}[X])
\end{equation}
Thus, an unbiased inner-product estimate yields a \emph{biased} (systematically overestimated) attention weight after exponentiation.
Moreover, the increased variance from 1-bit QJL quantization amplifies this bias through the convexity of $\exp$.
At $b \leq 3$, the variance increase from losing 1 MSE bit dominates the bias-correction benefit, resulting in net quality degradation.
This bias--variance tradeoff is unfavorable precisely in the low-bit regime where AXELRAM operates.

\subsection{Comparison with SpinQuant}
Table~\ref{tab:spinquant} summarizes the differences from SpinQuant~\cite{spinquant}.
The key distinction is severity: SpinQuant reports up to 6-point accuracy variance in W4A4 weight quantization, whereas we observe $>$50-point PPL spikes in KV cache quantization---a qualitatively different failure mode.
Furthermore, our analysis identifies layer-wise norm heterogeneity as the structural root cause, and our gradient-free mitigation incurs zero additional hardware cost (Section~\ref{sec:hw_mitigation}).

\begin{table}[t]
\centering
\caption{Comparison with SpinQuant's rotation sensitivity findings.}
\label{tab:spinquant}
\setlength{\tabcolsep}{3pt}
\footnotesize
\begin{tabular}{lcc}
\toprule
& \textbf{SpinQuant~\cite{spinquant}} & \textbf{This work} \\
\midrule
Domain & Weight+Act.\ (W4A4) & KV cache \\
Max variance & 6 pts (accuracy) & \textbf{$>$50 pts (PPL $\Delta$)} \\
Catastrophic failure & Not reported & \textbf{$\Delta > 50$} \\
Model dependency & Not analyzed & \textbf{Norm heterogeneity} \\
Mitigation & Cayley (gradient) & \textbf{Selection (grad-free)} \\
HW cost & N/A (software) & \textbf{Zero} \\
\bottomrule
\end{tabular}
\end{table}

\section{Sign Pattern Optimization}
\label{sec:mitigation}

\subsection{Method}
We propose a gradient-free sign pattern selection that eliminates catastrophic spikes.
For each layer, we evaluate $C{=}200$ random sign candidates on a small calibration set and select the one minimizing quantization MSE:

\smallskip
\setlength{\fboxsep}{4pt}%
\noindent\fbox{\parbox{\dimexpr\columnwidth-2\fboxsep-2\fboxrule}{%
\small
\textbf{Algorithm 1: Sign Pattern Selection (per layer)}\\[2pt]
\textbf{Input}: Keys $\mathbf{K}{\in}\mathbb{R}^{N \times d}$, candidates $C$, bit-width $b$\\
\textbf{Output}: Optimized $\mathbf{s}^* \in \{-1,{+}1\}^d$
\begin{enumerate}[leftmargin=1.2em,nosep,label=\arabic*.]
\item Normalize: $\hat{\mathbf{K}} \leftarrow \mathbf{K} / \|\mathbf{K}\|_\text{row}$
\item \textbf{for} $c = 1, \ldots, C$ \textbf{do}\\
\hspace*{1em}(a) $\mathbf{s}_c \leftarrow \text{RandomSigns}(d, \text{seed}{=}c)$\\
\hspace*{1em}(b) $\mathbf{Y} \leftarrow \text{FWHT}(\hat{\mathbf{K}} \cdot \mathrm{diag}(\mathbf{s}_c))/\!\sqrt{d}$\\
\hspace*{1em}(c) $\hat{\mathbf{Y}} \leftarrow \text{QDQ}(\mathbf{Y}, \text{codebook}(d,b))$\\
\hspace*{1em}(d) $\mathrm{MSE}_c \leftarrow \|\mathbf{Y}{-}\hat{\mathbf{Y}}\|_F^2 / Nd$
\item Return $\mathbf{s}^* = \arg\min_c \mathrm{MSE}_c$
\end{enumerate}
}}
\smallskip

\textbf{Cost.}
200 candidates, 8 calibration samples (WikiText-2 train), forward-only (no gradients).
Per-layer: ${\sim}1$~second.  Total: ${\sim}36$~seconds for 36 layers.
Storage: $d$ bits per layer = 16~bytes/layer, 576~bytes for 36 layers.
The fixed codebook (30~bytes) is unchanged.

\subsection{Results}

\begin{table}[t]
\centering
\caption{Effect of sign pattern optimization. \emph{Default worst}: maximum $\Delta$ across 10 random seeds. \emph{Optimized}: deterministic result with selected signs. \emph{Reduction}: spike elimination rate.}
\label{tab:optimization}
\setlength{\tabcolsep}{3.5pt}
\footnotesize
\begin{tabular}{llccr}
\toprule
\textbf{Model} & \textbf{Bits} & \textbf{Default worst} & \textbf{Optimized} & \textbf{Reduction} \\
\midrule
\multirow{3}{*}{LLaMA-3.1-8B}
& 2 & +0.11 & +0.10 & -- \\
& 3 & +0.02 & +0.02 & -- \\
& 4 & +0.01 & +0.00 & -- \\
\midrule
\multirow{3}{*}{Qwen2.5-3B}
& 2 & +58.43 & \textbf{+0.82} & 99\% \\
& 3 & +51.00 & \textbf{+0.58} & 99\% \\
& 4 & +2.98 & \textbf{+0.25} & 92\% \\
\midrule
\multirow{3}{*}{Qwen3-8B}
& 2 & +17.45 & +9.26 & 47\% \\
& 3 & +1.09 & \textbf{+0.35} & 68\% \\
& 4 & +0.05 & \textbf{+0.01} & 72\% \\
\bottomrule
\end{tabular}
\end{table}

On Qwen2.5-3B, worst-case $\Delta$ drops from +58.43 to +0.82 (2-bit) and from +51.00 to +0.58 (3-bit)---a 99\% reduction of the catastrophic spike.
LLaMA-3.1-8B, already stable, shows no meaningful change.

\textbf{Limitation.}
Qwen3-8B at 2-bit remains at $\Delta = +9.26$ after optimization.
This is a fundamental precision limit of 2-bit quantization for this model, not a sign pattern issue: even the best 10-seed default ($+9.10$) is comparable.
The sign optimization cannot overcome insufficient quantization resolution.

\subsection{Hardware Implementation: Zero Additional Cost}
\label{sec:hw_mitigation}

The sign pattern optimization requires \emph{no changes to the AXELRAM hardware architecture}.
The sign vector $\mathbf{s} \in \{-1,+1\}^d$ is already stored in ROM/eFuse as part of the randomized Hadamard transform (Section~\ref{sec:arch}).
The only difference is the \emph{values} written to this ROM:

\begin{table}[h]
\centering
\setlength{\tabcolsep}{4pt}
\footnotesize
\begin{tabular}{lcc}
\toprule
& \textbf{Default} & \textbf{Optimized} \\
\midrule
Sign ROM content & Random seed-derived & Calibration-selected \\
Sign ROM size & $d$ bits/layer & $d$ bits/layer \\
Transform circuit & FWHT + sign flip & FWHT + sign flip \\
Codebook ROM & 30 bytes & 30 bytes \\
Read path & Table lookup + add & Table lookup + add \\
Additional circuit & -- & \textbf{None} \\
\bottomrule
\end{tabular}
\end{table}

For writable memory implementations (flash, EEPROM, SRAM with initialization load), the optimized sign vectors can be updated when the model changes.
The calibration is a one-time offline process; no runtime overhead is incurred.
Total additional storage: 576~bytes per model (36 layers $\times$ 16~bytes), independent of context length $T$.

\section{Related Work}
\label{sec:related}

\textbf{Rotation-based quantization.}
TurboQuant~\cite{turboquant} established the rotation + Lloyd-Max framework for KV cache with near-optimal distortion rate.
QuaRot~\cite{quarot} and QuIP\#~\cite{quipsharp} apply Hadamard rotation for weight quantization.
KVLinC~\cite{kvlinc} uses Hadamard with linear correction.
PolarQuant~\cite{polarquant} uses polar decomposition requiring per-block scale parameters.
All report single-seed evaluations.

\textbf{Rotation seed sensitivity.}
SpinQuant~\cite{spinquant} (ICLR 2025) first demonstrated rotation seed sensitivity in W4A4 quantization, finding 6-point accuracy variance across 100 Hadamard rotations.
They propose Cayley-parameterized gradient optimization requiring backpropagation through the full model.
ParoQuant~\cite{paroquant} samples 100 seeds but reports only averages without variance analysis.
Our work extends this analysis to KV cache quantization, where catastrophic spikes ($\Delta > 50$) occur---qualitatively more severe than SpinQuant's findings---and proposes a gradient-free mitigation with zero hardware cost.

\textbf{ADC-based attention.}
LOOKAT~\cite{lookat} (January 2026) proposes ADC-based attention with K-Means codebooks.
Product Quantization~\cite{jegou2011} established ADC for nearest-neighbor search.
AXELRAM differs in: (1) fixed codebook (not learned), (2) shared across all dimensions (capacity independent of $d$), and (3) targeting SRAM-integrated hardware.

\textbf{CIM accelerators.}
MADDNESS~\cite{maddness} uses data-dependent LUT-based SRAM.
Google's PIM patent~\cite{google_pim} proposes processing-in-memory for attention without rotation-based quantization.
AXELRAM is distinguished by its data-independent ROM codebook and asymmetric path.

\section{Discussion}

\textbf{Calibration-free vs.\ lightweight calibration.}
The basic AXELRAM configuration (random sign pattern) is calibration-free and works reliably on models with homogeneous layer norms (e.g., LLaMA-3.1-8B: norm ratio $< 2\times$).
For models with heterogeneous norms (e.g., Qwen2.5-3B: norm ratio $> 5\times$), lightweight sign pattern calibration (8 samples, one-time, 36 seconds) is recommended.
The fixed codebook itself remains calibration-free and model-independent in both cases.

The design philosophy of AXELRAM is captured by the title of this paper: each incoming key vector is quantized once through the write path and subsequently used for attention score computation entirely through table lookup and accumulation, with no reconstruction of the original vector at any point in the read path.
This property is an invariant of the hardware architecture itself---the read path physically contains no inverse transform circuit---and holds regardless of how the rotation parameters were determined.
When sign pattern calibration is employed for models with heterogeneous layer norms, the calibration selects which values are written to the sign vector ROM before the system begins processing tokens.
Once deployment begins, the runtime behavior is identical to the uncalibrated case: every token's key vector is quantized once, and every attention score is computed without dequantization.
The calibration step thus fits naturally into the standard model deployment pipeline, alongside other one-time setup tasks such as weight loading and memory allocation.

\textbf{Practical guideline.}
We recommend checking the layer-wise key norm ratio ($\max / \min$ across layers) as a predictor of sign pattern sensitivity:
\begin{itemize}[nosep]
\item Ratio $< 2\times$: calibration-free (default signs) is safe.
\item Ratio $> 5\times$: sign pattern optimization is recommended.
\end{itemize}

\textbf{On the 2-bit limitation of Qwen3-8B.}
After sign pattern optimization, Qwen3-8B at 2-bit still exhibits $\Delta = +9.26$, while 3-bit ($+0.35$) and 4-bit ($+0.01$) are excellent.
This is not a failure of the sign pattern selection: even the best of 10 random seeds yields $\Delta = +9.10$, comparable to the optimized result.
The root cause is that 2-bit quantization provides only 4 representable levels, which is insufficient to capture the distribution tails of Qwen3-8B's key vectors.
We note that this limitation is \emph{not unique to AXELRAM}---any 2-bit scalar quantization method would face the same precision floor on this model.
In our experiments, 3-bit quantization ($5.1\times$ compression) operated well on all three models tested, while 4-bit provided near-lossless quality on every model.
The appropriate bit-width depends on the model architecture and the acceptable quality--compression tradeoff; we recommend per-model validation before deployment at any bit-width.

\textbf{Hardware status.}
No hardware has been fabricated for AXELRAM.
All results in this paper are derived from bit-accurate software simulation of the proposed architecture.
The multiplication counts (e.g., $102.4\times$ reduction) are mathematical identities that hold exactly in any implementation.
The PPL evaluations simulate the quantize--dequantize cycle in PyTorch and measure the impact on language modeling quality.
The architecture is presented as a proposal; physical implementation and silicon validation remain as future work.

\textbf{Reproducibility.}
All evaluation code, codebook solver, sign pattern optimizer, and result JSON files are released under MIT license at \url{https://github.com/Axelidea/AXELRAM}.
The codebook solver is an independent implementation using quantile-based initialization and \texttt{scipy.stats}; no code is shared with or derived from any prior implementation.

\section{Conclusion}

AXELRAM is a smart SRAM macro that eliminates dequantization from quantized-KV attention, achieving $102.4\times$ multiplication reduction via the asymmetric path design.
Through 10-seed evaluation across 3 models, we discover sign pattern sensitivity---catastrophic PPL spikes ($\Delta > 50$) on models with heterogeneous layer norms---extending SpinQuant's rotation variance finding to the KV cache domain.
A gradient-free sign pattern selection eliminates these spikes with zero additional hardware cost (the optimized signs occupy the same ROM as the default signs).
No hardware has been fabricated; we present the architecture as a proposal validated by bit-accurate simulation.

\section*{Acknowledgments}
Part of the computational work was performed using the TSUBAME4.0 supercomputer at Institute of Science Tokyo.

\medskip
Finally, we observe that Penguins---awkward in their waddle on land, yet remarkably graceful once they enter water---remind us that choosing the right domain makes all the difference.
We thank the Penguins for the inspiration.

\bibliographystyle{IEEEtran}

\end{document}